\documentclass[runningheads]{llncs}
\usepackage[T1]{fontenc}
\usepackage{amsmath}
\usepackage{graphicx}
\usepackage{amssymb}
\usepackage{algorithm}
\usepackage{algpseudocode}
\usepackage{tikz}
\usetikzlibrary{arrows.meta, positioning}

\usepackage{fixme}
\usepackage{cancel}
\usepackage{url}
\usepackage{booktabs}
\usetikzlibrary{arrows.meta, positioning}
\fxsetup{status=draft} % <====== add this line
  
\usetikzlibrary{shapes.geometric}
\usepackage{graphicx}
\usepackage{subcaption}
 
\usetikzlibrary{positioning}

\tikzset{
	box/.style={
		draw,
		rectangle,
		rounded corners,
		align=left,
		font=\small,
		inner sep=4pt,
		fill=green!15
	}
}

\begin{document}
   \title{On Solving the Multiple Variable Gapped Longest Common Subsequence Problem}
	\titlerunning{On Solving the MVGLCSP}
	
	\author{
		Marko Djukanović\inst{1,2, 6}  \and
		Nikola Balaban\inst{2}   \and
		Christian Blum\inst{3}  \and 
		Aleksandar Kartelj\inst{4}  \and
		Sašo Džeroski\inst{5}  \and
		Žiga Zebec\inst{6}
	}
	\authorrunning{Djukanovic et al.}
	\institute{University of Nova Gorica, Nova Gorica, Slovenia \\ \email{marko.dukanovic@ung.si} \\ \and Faculty of Natural Sciences and Mathematics, University of Banja Luka, Banja Luka, Bosnia and Herzegovina \\
		\email{nikola.balaban@student.pmf.unibl.org}, \email{marko.djukanovic@pmf.unibl.org} \and
		Artificial Intelligence Research Institute (IIIA-CSIC), Barcelona, Spain \\
		\email{christian.blum@iia.csic.es} \and 
		Faculty of Mathematics, University of Belgrade, Belgrade, Serbia \\ \email{kartelj@matf.rs} \\ \and
		Jožef Stefan Institute, Ljubljana, Slovenia \ \\ \email{saso.dzeroski@ijs.si}  \\
		\and
		Institute of Information Sciences (IZUM), Maribor, Slovenia \\
		\email{ziga.zebec@izum.si} \vspace{-0.7cm} \\
	}
	\maketitle              % typeset the header of the contribution

\begin{abstract}	\vspace{0.15cm}
    This paper addresses the Variable Gapped Longest Common Subsequence (VGLCS) problem, a generalization of the classical LCS problem involving flexible gap constraints between consecutive solutions' characters. The problem arises in molecular sequence comparison, where structural distance constraints between residues must be respected, and in time-series analysis where events are required to occur within specified temporal delays. We propose a search framework based on the root-based state graph representation, in which the state space comprises a generally large number of rooted state subgraphs. To cope with the resulting combinatorial explosion, an iterative beam search strategy is employed, dynamically maintaining a global pool of promising candidate root nodes, enabling effective control of diversification across iterations. To exploit the search for high-quality solutions, several known heuristics  from the LCS literature are utilized into the standalone beam search procedure. To the best of our knowledge, this is the first comprehensive computational study on the VGLCS problem comprising 320 synthetic instances with up to 10 input sequences and up to 500 characters. Experimental results show robustness of the designed approach over the baseline beam search in comparable runtimes. %Furthermore, for the special case with two input strings, the proposed method achieves great performance compared to the specialized dynamic programming algorithms on small-to-medium-sized instances.
\end{abstract}
\section{Introduction}
	
The \textit{Longest Common Subsequence Problem} (LCSP)~\cite{DJUKANOVIC2020106499,bergroth2000survey} is a well-known combinatorial optimization problem with numerous applications in bioinformatics and computational biology, playing a fundamental role in the analysis and comparison of molecular sequences. Given an arbitrarily large set of input sequences $
S=\{s_1,\ldots,s_m\}, m \in \mathbb{N}$  over an alphabet $\Sigma$, the aim is to identify the longest possible subsequence that is common to all sequences $s_i \in S$. Over the past decades, a variety of practically motivated extensions of the LCSP have been proposed to better capture structural, biological, or application-specific requirements of real-world sequences. Notable variants include constrained, arc-preserving, and repetition-free longest common subsequence problems~\cite{jiang2004longest,farhana2015constrained,adi2010repetition}, among others. In this work, we focus on yet another practical variant, the \textit{Variable Gapped LCS Problem} (VGLCSP),  originally introduced for the $m=2$ case in~\cite{penga2011longest}, later investigated from a theoretical perspective in~\cite{adamson2023longest}. The VGLCSP extends the classical LCSP by incorporating flexible distance constraints (referred to as \emph{gaps}) between consecutive symbols of the resulting subsequences and their respective positions in the input sequences. Unlike fixed-gap models, these gap limits are allowed to vary along the input sequences, thereby offering increased modeling flexibility.\\
Formally, for each input sequence $s_i \in S$, its gap constraint is defined as a function over the  positions of the sequence, i.e., $
G_i \colon \{1,\ldots,|s_i|\} \mapsto \mathbb{N}.$ Further, if the characters of a solution $s$ occur at positions
$i^1_i < \cdots < i^{|s|}_i$ in $s_i$, the gap constraint is satisfied iff $
i^x_i - i^{x-1}_i \leq G_i[i^x_i] + 1, \quad x = 2,\ldots,|s|.$ A common subsequence $s$ is considered \emph{feasible} if the corresponding gap constraints are satisfied simultaneously for all sequences in $S$. As an example of a VGLCSP instance, given is a set  $S=\{s_1=\texttt{ABCA},   s_2=\texttt{ACAB} \}$ with two gap constraints $G_1(\{1,\ldots, 4\}) \mapsto \textbf{1}= G_2(\{1, \ldots, 4\})$. The sequence $s=\texttt{ACA}$ is a feasible solution because it is a common subsequence for both input strings, where its corresponding characters appear at positions 1, 3, and 4 in $s_1$ and at the positions 1, 2, and 3 in $s_2$. As the difference between each consecutive pair of the indices of these positions is less than or equal to $1+1=2$, $s$ represents a feasible sequence of the tackled problem.  On the other hand, sequence $s=\texttt{AB}$ represents a common subsequence, but not fulfilling all gap constraints because the corresponding characters appear at positions 1 and 2 in $s_1$ and at positions 1 and 4 in $s_2$. Obviously, $4-1=3>2=1+1$, thus violating the second gap constraint. \\ 
The VGLCSP provides a flexible and realistic model for sequence comparison, particularly suiting to applications in DNA and protein analysis where variable structural distances between (active) residues must be respected. Beyond bioinformatics, the problem also arises in time-series analysis, especially in settings where events are required to occur within specified temporal delays~\cite{lainscsek2015delay}.\\
Although several dynamic programming approaches exist for the $m=2$ case~\cite{penga2011longest}, they are computationally prohibitive when expanded for larger $m$. To the best of our knowledge, no effective exact or approximate approaches are designed for the generalized VGLCSP with arbitrary $m$, clearly an $\mathcal{NP}$-hard problem as the generalization of LCS problem with an arbitrarily large set of input strings. \\
\vspace{-0.7cm}
\paragraph{Preliminaries.}	Let $|s|$ denote the length of a sequence $s$, and let $s[i]$ refers to the character of $s$ at position $i$, where indexing starts at $i = 1$. The substring of $s$ that begins at position $i$ and ends at position $j$ is denoted by $s[i,j]$. If $i = j$, this corresponds to the single character $s[i]$; if $j < i$, then $s[i,j] = \varepsilon$, where $\varepsilon$ denotes the empty string.	The number of occurrences of $a \in \Sigma$ in $s$ is denoted by $|s|_a$.	We denote by $S = \{s_1, \ldots, s_m\}$ the set of input sequences, and by $S^{rev} = \{s_1^{rev}, \ldots, s_m^{rev}\}$, the set of reverse strings.  
Let $m \in \mathbb{N}$ be the number of input sequences (and, correspondingly, the number of gap constraints), and $n$ denotes the length of the longest sequence in $S$.  Given a positional vector $\mathbf{p}^L \in \mathbb{N}^m$,  the subproblem related to these positions is given by $S[\mathbf{p}^L]=\{ s_i[\mathbf{p}^L_i, |s_i|] \mid i=1, \ldots, m\}$.  Last but not least, for a gap constraint $G_i$, its reverse constraint is denoted by  $G_i^{rev}$, given by $G_i^{rev}(j):=G_i(|s_i|-j+1)$, for each $j=1, \ldots, |s_i|$.  \\ 

	 The remainder of the paper is organized as follows. Section~\ref{sec: state_graph} introduces the rooted graph representation of the problem. In Section~\ref{sec: imsbs}, we  design our main methodological contribution, the iterative multi-source beam search approach. Section~\ref{sec: experiments} reports the computational results, comparing the proposed method with a baseline beam search and an iterative greedy heuristic. %, but also with several specially designed  approaches for two input sequences. 
	Finally, Section~\ref{sec: conclusions} concludes the paper and outlines directions for future research.

	\section{Rooted State Space Formulation}\label{sec: state_graph}

Motivated by the state-space representation for the LCSP introduced in~\cite{DJUKANOVIC2020106499}, we design a \emph{rooted state graph} model for the VGLCSP. Each state represents one or more feasible partial solutions characterized by a vector of positions induced by the input sequences and by the current subsequence length. 

Formally, a partial subsequence $s^v$ induces a state node $
v = (\mathbf{p}^{L,v},\, l^v),$ 
where $\mathbf{p}^{L,v} = (p^{L,v}_1,\ldots,p^{L,v}_m)$ and the following conditions hold:
(i) for each sequence $s_i \in S$, $p^{L,v}_i - 1$ is the smallest position such that $s^v$ is a subsequence of $s_i[1,\, p^{L,v}_i - 1]$;
(ii) $l^v = |s^v|$ denotes the length of the partial subsequence; and
(iii) all gap constraints $G_i$ are \textit{satisfied} for $s^v$ with respect to each input sequence $s_i$. A direct arc $\alpha = (v_1, v_2)$, labeled with $\texttt{lett}(\alpha) = a \in \Sigma$, exists if $
l^{v_2} = l^{v_1} + 1
\quad \text{and} \quad
s^{v_2} = s^{v_1} \cdot a$; 
that is, state $v_2$ corresponds to extending a partial solution of $v_1$ by (appending) character $a$, respecting all gap constraints. 

To expand a state $v$, those characters $a \in \Sigma$ that occur in each suffix  $s_i[p^{L,v}_i, |s_i|]$, $i = 1,\ldots,m$, are first detected. For each such $a \in \Sigma$, we identify the smallest feasible positions $p^{L,a}_i \ge p^{L,v}_i$ such that
$
s_i[p^{L,a}_i] = a
\quad \text{and} \quad
p^{L,a}_i - p^{L,v}_i \le G_i(p^{L,a}_i), i = 1,\ldots,m.
$ 
In that way, a child state
$w = (\mathbf{p}^{L,a} + \mathbf{1},\, l^v + 1)$
is created, unless it is dominated by another child state according to the dominance criteria defined later in this section. To efficiently compute children and their associated position vectors, we employ a preprocessing data structure denoted by \texttt{Succ}: a three-dimensional integer array indexed by $(i,j, a)$, where $i$ refers to the input sequence $s_i$, $1 \le j \le |s_i|$ denotes a position within that sequence, and $a \in \Sigma$. The value $q = \texttt{Succ}[i,j,a]$
represents the smallest index $q \ge j$ such that $s_i[q] = a$ so that the gap constraint $G_i$ is satisfied at position $q$, i.e.,  $q - j \le G_i(q).$ 
If no such position exists, $q=-1$ is assigned.  Goal states are those that cannot be further expanded.    The trivial root state is given by $ r = ((1,\ldots,1),\, 0)$ as in-going edges to this node are not possible,   representing an empty partial solution like in the case of the LCS problem. However, this is not the only one root node, in general. The state graph of a problem can have many (possibly exponentially) root nodes, detecting the essential decision in selecting leading letter for future  expansions. %,  induced from its position of occurrence defined by the 
Note that if match $u=(i^1_1, \ldots, i^1_m)$ represents a starting position,   the gap constraints do not apply to the corresponding letter of this or any other  rooted match. 

\begin{figure}[h!] \vspace{-0.6cm}
	\centering
	\scalebox{0.35}{

		\begin{tikzpicture}[
			state/.style={circle, draw=blue!50, fill=blue!10, thick, minimum size=8mm},
			edge/.style={-Latex, thick},
			node distance=2cm and 2cm
			]
			
			% Čvorovi
			\node[state, fill=blue] (s0) {((1, 1),0)};
			\node[state, below left=of s0] (s1) {((2, 2),1)};
			\node[state, below right=of s0] (s2) {((4,3),2)};
			%\node[state, below=of s2] (s3) {(4,3)};
			\node[state, below=of s2, fill=gray] (s4) {((5,4),3)}; % krajnje, ali ne koristi se ovde
			\node[state, left=of s4, fill=white] (s5) {((3, 5), 2)}; % krajnje, ali ne koristi se ovde
			
			% Right of root ((1,1),0)
			\node[box, right=12mm of s0] (b0) {
				$s_1=\text{{A}BCA}$\\
				$s_2=\text{{A}CAB}$\\
				$s^v=\varepsilon$
			};
			
			% Left of ((2,2),1)
			\node[box, left=14mm of s1] (b1) {
				$s_1=\text{\textcolor{gray}{A}BCA}$\\
				$s_2=\text{\textcolor{gray}{A}CAB}$\\
				$s^v=\text{A}$
			};
			
			% Right of ((4,3),2)
			\node[box, right=12mm of s2] (b2) {
				$s_1=\text{\textcolor{gray}{ABC}A}$\\
				$s_2=\text{\textcolor{gray}{AC}AB}$\\
				$s^v=\text{AC}$
			};
			
			% Right of ((5,4),3)
			\node[box, right=42mm of s5] (b3) {
				$s_1=\text{\textcolor{gray}{ABCA}}$\\
				$s_2=\text{\textcolor{gray}{ACAB}}$\\
				$s^v=\text{ACA}$
			};

			% Grane (samo validne poklapanja)
			\draw[edge] (s0) -- node[above] {\small A} (s1);
			%\draw[edge] (s0) -- node[above] {\small A} (s2);
			\draw[edge] (s1) -- node[above] {\small C} (s2);
			%\draw[edge] (s2) -- node[right] {\small A} (s3);
			%\draw[edge] (s3) -- node[right] {\texttt{A}} (s4);
			\draw[edge,  scale=2.2, dashed, red] (s1) -- node[right, fill=lightgray] {$\cancel{\texttt{B}}$} (s5);
			
			\draw[edge,  scale=2.2, dashed, red] (s1) -- node[right, fill=lightgray] {$\cancel{\texttt{A}}$} (s4);
			% Oznake
			\node[above=0.3cm of s0] {\textbf{Root node}};
			%\node[below=1.5cm of s3] {\small \textbf{Krajnje stanje}};
			\draw[edge] (s2) -- node[right] {\texttt{A}} (s4);
			
	\end{tikzpicture}}
	
	\caption{  $\textsc{RootedSubspace}(u=((1, 1), 0))$  between   \texttt{\fbox{A}BCA} and  \texttt{\fbox{A}CAB}, assuming $G_1=G_2=\textbf{1}$. A final/terminal state is displayed in gray. The white-filled node is a valid LCS node, but not an VGLCS node---the second gap constraint is violated. The longest solution generated from this subspace is \texttt{ACA}. % \fxnote{Add VGLCS example}
	}
	\vspace{-0.7cm}
	\label{fig:vgcs-grafstanja}
\end{figure}
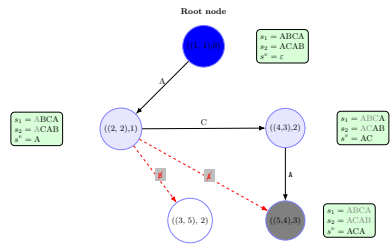

This motivate us to introduce the state space subgraph rooted with match $\mathbf{p}^{L, u}$, denoted by $\textsc{RootedSubgraph}(u=(\mathbf{p}^{L, u}, 0))$, where $u$ is induced by empty partial solution. The empty solution  is initially expanded by the letter $l$ matching $\mathbf{p}^{L, u}$ as we require to generate all feasible solutions having $l$ for the leading letter.  
In essence,  this subgraph consists of all nodes (and corresponding transitions)  feasibly reached via node $(p^{L,u}+\textbf{1}, 1)$. These nodes are associated to partial solutions with a leading letter to match position $\mathbf{p}^{L, u}$. We recursively repeat the procedure of expanding solutions by one letter in all possible ways, until no feasible expansions are possible.  As discussed in the following section, the structure of this subspace heavily depends on the choice for root state $u$ from which the search is initiated. We illustrate  the space of \textsc{RootedSubgraph}($u=((1, 1), 0)$),  depicted in Fig.~\ref{fig:vgcs-grafstanja}. Note that here the position $(1,1)$ correspond to the match of letter \texttt{A} associated to $u=((1, 1), 0)$ which is indeed a root node. 

   % U nekom delu LaTeX dokumenta:

\textit{Multi-Source Beam Search.}  
Unlike the classical LCSP, the VGLCSP may exhibit multiple, potentially exponentially many, \emph{disconnected root components} in the full state space.  Consequently, a search strategy that expands states from a single initial root node may fail to reach optimal solutions. 
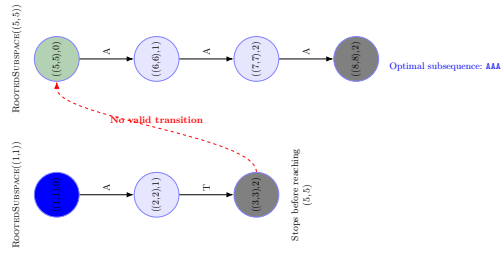
\begin{figure}[!ht] 
	\centering
	\scalebox{0.40}{
		\begin{tikzpicture}[
			rotate=90,
			transform shape,
			state/.style={circle, draw=blue!50, fill=blue!10, thick, minimum size=8mm},
			edge/.style={-Latex, thick},
			invalid/.style={-Latex, thick, dashed, red},
			node distance=1.8cm
			]
			% ---------------- LEFT COMPONENT ----------------
			\node[state, fill=blue] (r1) at (0,0) {((1,1),0)};
			\node[state, below=of r1] (a1) {((2,2),1)};
			\node[state, below=of a1, fill=gray] (a2) {((3,3),2)};
			
			\draw[edge] (r1) -- node[right] {\small A} (a1);
			\draw[edge] (a1) -- node[right] {\small T} (a2);
			
			\node[above=0.3cm of r1] {\textsc{RootedSubspace}((1,1))};
			\node[below=0.3cm of a2, align=center] {\small Stops before reaching \\ $(5,5)$};
			
			% ---------------- RIGHT COMPONENT ----------------
			\node[state, fill=green!40!black!30, right=3cm of r1] (r2) {((5,5),0)};
			\node[state, below=of r2] (b1) {((6,6),1)};
			\node[state, below=of b1] (b2) {((7,7),2)};
			\node[state, below=of b2, fill=gray] (b3) {((8,8),2)};        
			\draw[edge] (r2) -- node[right] {\small A} (b1);
			\draw[edge] (b1) -- node[right] {\small A} (b2);
			\draw[edge] (b2) -- node[right] {\small A} (b3);
			
			\node[above=0.3cm of r2] {{$\textsc{RootedSubspace}((5,5))$}};
			\node[below=2.2cm of b3, rotate=-90] {\small \textcolor{blue}{Optimal subsequence: \texttt{AAA}}};
			
			% ---------------- DISCONNECTION INDICATOR ----------------
			\draw[invalid] (a2.east) .. controls +(1.2,0) and +(-1.2,0) .. 
			node[above, rotate=-90] {\small \textbf{No valid transition}} (r2.west);
			
		\end{tikzpicture}
	}
	\caption{Root state space subgraphs: disconnected components.} \vspace{-0.5cm}
	\label{fig:disconected-components}
\end{figure}

As an illustrative example given in Fig.~\ref{fig:disconected-components}, consider the sequence set $S = \{ s_1 = \texttt{ATGGAAA},\; s_2 = \texttt{ATCCAAA} \},$
with  $G_{s_1} = G_{s_2} = {1}$. In this instance, any state with position vector $\mathbf{p}^L = (5,5)$ is not reachable from the initial state $((1,1),0)$ by a series of direct transitions. As a result, the optimal common subsequence \texttt{AAA} is unreachable from $((1,1),0)$  when exploring its subgraph. To address this issue, we define the complete state space of a VGLCSP instance by  
$\bigcup_{v \in \mathcal{R}} \textsc{RootedSubspace}((\mathbf{q}^{L, v},0))$,  
where $\mathcal{R}$ denotes the set of all root states,  
which cannot be reached from any other state by a series of direct transitions. However, explicitly enumerating all such root states is computationally prohibitive and in generally requires $O(n^m)$ time, thus remains infeasible for instances with large $m$ or $n$.  This structural characteristic of the problem itself motivates the methodology entitled \textit{Iterative Multi-Source Beam Search} (IMSBS) approach,  specifically designed to dynamically explore multiple promising rooted subregions of the state space by systematically moving exploration from one subgraph to another one. By iteratively identifying and expanding a set of candidate root states, IMSBS effectively mitigates the disconnection effects induced by gap constraints, enabling the discovery of high-quality solutions that would otherwise remain inaccessible to single-source search methods.

\section{Iterative Multi-source Beam Search Framework}\label{sec: imsbs}
%	Before presenting the main details on the Iterative Multi-source Beam Search (IMSBS) approach, we give a brief overview over a general beam search metaheuristic approach as its crucual component. 
	
	\textit{Beam search} (BS) is a well-established tree-search-based metaheuristic that performs in a breadth-first-search manner. Its search is controlled by a beam width parameter $\beta$---specifying the maximum number of nodes retained for expansion at each level---and by  heuristic guidance $h$--- which guides the selection of the most promising candidates for further expansion. Initially, the procedure receives a set of nodes as the initial beam. After being evaluated, up to $\beta$ most promising nodes are kept for creating a beam of the subsequent level. The procedure repeats until the beam is empty, returning the longest partial solution.

In the IMSBS, the beam search component serves to intensify the search process. As the search space of a VGLCSP instance is represented by the set of root nodes $\mathcal{R}$, the BS procedure is in charge of efficiently exploring various  regions \textsc{RootedSubspace}($(\mathbf{p}^L, 0)$) induced by corresponding root nodes. Note that one could here employ an exact technique (such as A$^*$ or BFS); however, solving the subproblem itself could be a resource-demanding task, as hard as the original problem itself. Thus, we opted to utilize a more robust BS technique.

%The \textit{Iterative Multi-Source Beam Search} framework, standalone BS procedure is executed iteratively, each time initialized with a new beam $\mathcal{L}$ consisting of several promising candidates root nodes. These nodes are dynamically selected from a global pool $\mathcal{R}$, whose role is to promote diversification across different regions of the state space. Each BS execution thus explores the neighborhood of multiple starting points rather than relying on a single root $r$.

%The beam search itself employs several heuristic components to intensify the search toward high-quality feasible solutions. These heuristics include estimates based on letter frequencies, minimal residual substring lengths, and probability-weighted bounds, which are described in detail in the subsequent sections. \\
The main core of the IMSBS framework is summarized as follows. % in Algorithm~\ref{alg:imbs}. 
Initially, the global pool of candidate root nodes is initialized to an empty set $\mathcal{R}:=\{\}$ and the current best solution to $s_{\mathrm{best}}:= \varepsilon$. First, for each $a \in \Sigma$, the closest matches $\mathbf{p}^{L, a}$ from the leading positions of each input sequence are detected. Each such (non-dominated) match generates one root node $(\mathbf{p}^{L, a}, 0)$, which has been subsequently added to $\mathcal{R}$.  The core iterations of IMSBS have been executed until either  $\mathcal{R}$ becomes empty, a predefined maximum number of iterations has reached, or allowed time limit has exceeded. At each iteration, a fixed number (\emph{sources\_from\_R}) of most promising candidate root nodes according to a heuristic $h'$ is taken from $\mathcal{R}$, and subsequently stored in a temporary beam $\mathcal{L}$.  These nodes are refined, executing a single backward beam search procedure for each $w \in \mathcal{L}$ as its starting node $(\mathbf{p}^w, 0)$. The procedure works similarly to the baseline BS,  expanding nodes according to the gap constraints but in the reversed manner, starting from the root positions at each sequence and moving the search (pointers) towards their leading positions. In essence, for each $w \in \mathcal{L}$, a beam search  %\textsc{RootedBeamSearch}
 is executed on the reversed input sequences $S^{rev}$ and the reversed gap constraints $(G_i^{rev})_{i=1}^m$ with the initial beam width $B=\{ ((|s_i|-p^w)_{i=1}^m, 0) \}$, beam size $\beta'$ and heuristic guidance $h'$.   %an initial beam consisting of the node $((|s_i|-p^w - 1)_{i=1}^m, 0)$. 
%In details, a letter $a \in \Sigma$ is a feasible extension of partial solution associated with node $v =(p^v, l^v)$ iff it appears in all $s_i[1, p^v_i], i=1, \ldots, m$ and for the largest indices $p^w_i <  p^u_i$  such that $s_i[p^w_i]=a$, it holds $ p^v_i - p^w_i  \leq G_i[p^v_i] + 1 $, for each $i=1, \ldots, m$. The child node is given by $w = (p^w - \textbf{1}, l^u +1 )$.
Assume the procedure has completed, returning a solution $s^w_{bwrd}$. In this way, the reverse sequence $s_{bwrd}^{rev}$  represents a feasible VGLCS solution and the partial solution without the last letter---which corresponds to the match $\textbf{p}^w$---is assigned to the refined node $ w \in \mathcal{L}$, and the updated node $w=(p^w, |(s^w_{bwrd})^{rev}|-1)$ is passed to the forward beam search procedure for the forward-pass beam search. Thus, the refined set $\mathcal{L}$ enters in as the initial beam, executing a BS parametrized with $\beta$ and $h$, following the subspace definition in Section~\ref{sec: state_graph}.  During this phase, all complete (non-expandable) nodes are stored in the list $V_{complete}$, which is, after termination,  returned along with the best found solution $s_{b}$. %, later verified for a new incumbent.  % Further, if a complete node from $V_{complete}$ corresponds to a solution that improves upon the current best, the solution becomes a new incumbent.   
Each complete node from $V_{complete}$  is then independently expanded---ignoring the gap constraints--- to generate a set of candidate root nodes. In details, for each complete node $v =(\mathbf{p}^v, l^v)$, 
candidate nodes  $r^{w, a} = (\mathbf{p}^{w, a}, 0)$ are generated for each letter $a \in \Sigma$, where $p^{w, a}_i$ represents the match position of that letter in $s_i$ which is closest from position $\textbf{p}^v_i$. Each such candidate node $(r^{w, a}, 0)$ is added to $\mathcal{R}$, given that it has not already been part of $\mathcal{R}$ during the previous iterations.  Once the BS procedure is terminated, returned solution if verified as a new incumbent, and  the next \textsc{Imsbs} iteration is initiated with a newly selected set of most promising root nodes from $\mathcal{R}$, until a termination condition is met. The pool $\mathcal{R}$ may be maintained as a priority queue implemented via a binary heap, to ensure efficient extraction of the most promising candidate root nodes. \\
The core advantages of the IMSBS approach over the baseline BS are: (i)  A provision of balance between complete beam search execution (local exploration of promising search subregions) and the generation of new candidate root nodes (global coverage of the search space); (ii) Prevention from suboptimal candidate root nodes is achieved by executing the backward beam search procedure.\\
\textit{Heuristic guidances.} Several heuristic guidances to guide search are utilized,  known from the LCS literature: (i)  \textit{``Look-ahead'' for the remaining length:} $\textrm{UB}_1(\mathbf{v}) = l^v + \min_{i = 1, \ldots, m} \left( |s_i| - p_i^{L, v} + 1 \right)$;
%	This is an estimate of the potential extension of the sequence up to the end of the shortest remaining sequence.
(ii)  \textit{Character Frequency Alignment:}
$\textrm{UB}_2(\mathbf{v}) =  l^v + \sum_{\sigma \in \Sigma} 
\min\left(  |s_1[p_1^{L, v}, |s_1|]|_{\sigma}, \ldots, |s_m[p_m^{L, v}, |s_m|]|_{\sigma} \right)$,  counting the maximum number of characters that can still match in the (optimal) solution based on the frequency of each character; %This score can be computed efficiently in \(O(m)\) time using preprocessing (i.e., constructing suitable data structures), see the details in~\cite{DJUKANOVIC2020106499}. 
(iii) \textit{A probability-based heuristic} guidance $h_{prob}$ by the Mousavi and Tabataba's matrices~\cite{mousavi2012improved}:  These probabilities are based on determining a probability of the event realization that a sequence $s$ of length $k$ is a subsequence of a (random) sequence of length $n$ over $\Sigma$.  %The probabilities for each  $i=1,\ldots, k$ and  $j=1, \ldots, n$, are pre-processed by a matrix $\mathcal{P}$ of probabilities with dimension $k\times n$ effectively generated by a dynamic programming. Assuming independence between input sequences,  the nodes/partial solutions within the same level of beam search are evaluated by determining the associated probabilities of their expansions by $\overline{k}$ letters (this value is heuristically evaluated, as proposed in~\cite{mousavi2012improved}). Denote that heuristic guidance   by $h_{\textrm{prob}}$. 
     More about these heuristics interested readers can found at~\cite{DJUKANOVIC2020106499,mousavi2012improved}. 
     %\item TODO: \fxnote{TODO: if we come up with some learning heuristic -- should left for the extended version -- journal-extendable.}
   %\end{itemize}

\section{Experimental Evaluation}\label{sec: experiments}
   
The performance of heuristic approaches designed for an arbitrary \( m \geq 2 \) are experimentally compared. Specifically, the following methods are included: (i) \textsc{Bs}: a baseline beam search, which performs a single iteration of IMSBS while employing a huge beam width; (ii) \textsc{Imsbs-greedy}: it sets a fixed  \( \beta = 1 \) %, as given in Algorithm~\ref{alg:imbs}.  
   employing a huge number of IMSBS iterations; (iii) \textsc{Imsbs}: a tuned IMSBS with an average runtime comparable to that of \textsc{Bs}.	 The \textsc{Imsbs} is implemented in Python~3.11 and evaluated on the VEGA HPC system hosted at IZUM, Maribor. The cluster consists of 960 compute nodes equipped with AMD EPYC 7H12 CPUs operating at 2.35\,GHz. All experiments were conducted in a single-threaded setting.  For each combination of  $n \in \{50, 100, 200, 500\}$, $m \in \{2, 3, 5, 10\}$, and $|\Sigma| \in \{2, 4\}$, 10 random problem instances are generated. First, $m$ sequences of equal length are generated uniformly at random. Then, the gap constraints for each instance are generated such that each value $G_i(j), j=1, \ldots, |s_i|, i=1, \ldots, m$ is  assigned uniformly from 
 $\mathcal{U}(\{ \lfloor 0.5 \cdot |\Sigma| \rfloor, \ldots, \lfloor 1.5 \cdot |\Sigma| \rfloor \})$.  Therefore, a total of \textit{320 problem instances} is generated. The benchmark set is denoted as \textsc{Random}, freely available at   
 \url{https://github.com/markodjukanovic90/NikolaBalabanDiplomski/tree/main/src/instance_i_generator}.
 
 \textit{Parameter tuning}. 
 %\fxnote{Came up to here!!}
 To tune the parameters of our algorithms, 
 we employed a basic grid search to more sensible parameters: beam width in the beam search forward pass ($\beta$), and heuristic guidance  $h \in \{\text{UB}_1, \text{UB}_2, h_{prob}\}$, comparing the average solution quality over all (320) problem instances.  Less sensible parameters are selected according to several preliminary runs.  %as further increasing the beam size does not significantly improve solution quality. The heuristic $h_{\text{prob}}$ provides slightly better guidance than $\textrm{UB}_2$ and significantly better than $\textrm{UB}_1$; see Fig.~\ref{fig:bs_tuning}.
 We fix ${{sources\_from\_R}} = 10$ and $\emph{beam\_iters} = 100$ (beam iterations in \textsc{Imsbs}), and %as good results were already achieved employing these numbers. Root nodes kept in the global set are prioritized according to 
  $h' = \textrm{UB}_2$.   For the baseline \textsc{Bs} approach, a high beam width $\beta = 10,000$ and $h = h_{\text{prob}}$ are utilized. %, which consistently improves IMSBS performance compared to using $\textrm{UB}_1$.  
  Remaining parameters of the \textsc{Imsbs} approach received the following values: $h=\text{UB}_2$ and $\beta=500$ to keep comparable average runtime between \textsc{Imsbs} and the baseline \textsc{Bs} approach. These values balance runtime relative to \textsc{Bs} while maintaining a reasonable solution quality.  Similarly, for the \textsc{Imsbs-greedy} approach with fixed $\beta = 1$ uses  $\emph{beam\_iters} = 10,000$, and all other parameters hold as in the case of tuned \textsc{Imsbs}.  
  
   \begin{figure}[!ht]
  	\vspace{-1.0cm}
  	\begin{subfigure}[t]{0.45\textwidth}
  		\centering
  		\includegraphics[width=\linewidth, height=90pt]{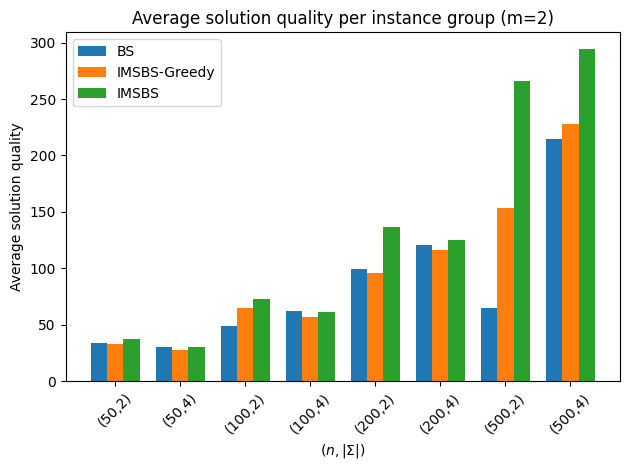}
  		\caption{$m=2$}
  		\label{fig:results_per_m2}
  	\end{subfigure}
  	~
  	\begin{subfigure}[t]{0.45\textwidth}
  		\centering
  		\includegraphics[width=\linewidth, height=90pt]{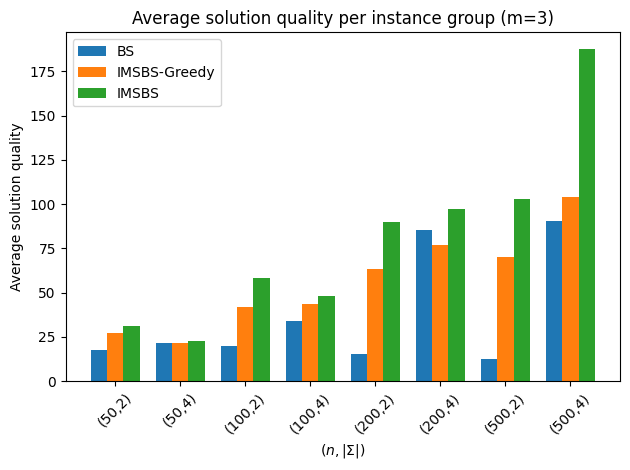}
  		\caption{$m=3$}
  		\label{fig:results_per_m3}
  	\end{subfigure}
  	
  	\centering
  	\begin{subfigure}[t]{0.45\textwidth}
  		\centering
  		\includegraphics[width=\linewidth, height=90pt]{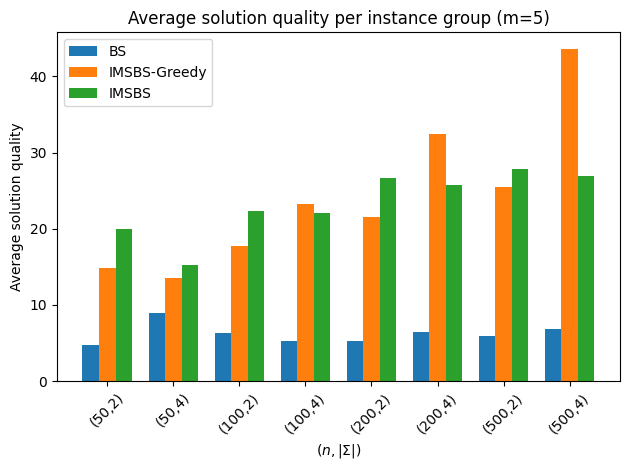}
  		\caption{$m=5$}
  		\label{fig:results_per_m5}
  	\end{subfigure} 
  	~
  	\begin{subfigure}[t]{0.45\textwidth}
  		\centering
  		\includegraphics[width=\linewidth, height=90pt]{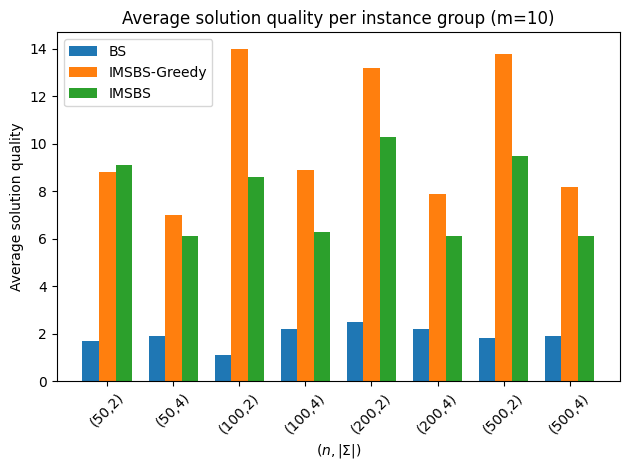}
  		\caption{$m=10$}
  		\label{fig:results_per_m10}
  	\end{subfigure}
  	\caption{Average solution quality of the three approaches.} \vspace{-0.5cm}
  	\label{fig:results_per_m}
  \end{figure}

  Numerical results for all three heuristic approaches are displayed in Fig.~\ref{fig:results_per_m}, reporting the average solution quality per instance group for each approach. Four plots are provided, one per each value of $m$. %Based on the numerical results, %presented in Table~\ref{tab:numerical_results_general_m}, 
  %the following conclusions are drawn.
  %in Table~\ref{tab:numerical_results_general_m}, which is organized into two main parts. 
  %The first three columns describe the instance characteristics: the number of sequences ($m$), the sequence length ($n$), and the alphabet size ($|\Sigma|$). 
  %The remaining columns report the performance of the three heuristic approaches: \textsc{Bs}, \textsc{Imsbs-greedy}, and \textsc{Imsbs}, each in its own block. For each algorithm, two performance indicators are provided: the average objective value $\overline{obj}$ and the average running time in seconds $\overline{t}$, computed over 10 instances per group.  
   The \textsc{Imsbs} approach produces the highest average solution quality in 21 out of 32 instance groups, surpassing the baseline \textsc{Bs} approach dramatically, indicating that the selection of the right root node is essential in achieving high-quality solutions.   \textsc{Bs} performs better in just one case. Concerning the \textsc{Imsbs-greedy}, it surpasses the other two approaches in 10 (out of 32) cases, mostly for the instances with larger $m=10$, having shorter final solutions where it turns out that frequently moving from one to another root and exploring various subregions of the search space is strategically crucial than just invest the time for strengthening the search around a solution employing a hige $\beta$.  Exploring additional root nodes improves the diversity of the search, which appears necessary in cases where expected solutions are short. This is exactly where  \textsc{Imsbs-greedy} fits well. 
\textsc{Bs} and \textsc{Imsbs} approaches are tuned to exhibit comparable average runtimes.  \textsc{Imsbs-greedy} has a larger average runtimes for the instances with larger $m$ due to a higher number of executed iterations ($beam\_iters = 10,000$ iterations allowed) which are not cheap when $m$ is large. Detailed numerical results with the accompanying plots can be found in the supplementary material at  \url{https://github.com/markodjukanovic90/VGLCS/blob/main/doc/EUROCAST-paper/supplementary.pdf}.  % in the other two approaches.  %(limited to $beam\_iters = 100$).
  	
  %	\item \textit{Other conclusions.}  Overall, \textsc{Imsbs} demonstrates greater robustness and stability as the number of input sequences increases in comparison to the baseline \textsc{Bs} approach which performance is gradually deteriorated. 
  	
  	%As illustrated in Figure~XX, both BS and IMSBS frequently attain near-optimal or optimal solutions for smaller $n$ and larger alphabets. The BS approach reaches at least 65\% of the optimum for the majority of instances, whereas IMSBS achieves 80\% or more. For approximately 50 out of the 60 instances, both methods achieve at least 90\% of the optimal solution length, as shown in Figure~XX. 

\section{Conclusions and Future Work}
\label{sec: conclusions}

This paper studied a generalized variable gapped LCS problem. It extends the classical LCS problem by constraining the allowable distances between matched characters, reflecting biological scenarios in which spatially closer nucleotides interact more strongly than distant ones.  
We proposed an Iterative Multi-Source Beam Search (IMSBS), which performs local exploration from promising root nodes involving a global strategy of candidate root-node selection based on previously collected search information. Experimental results indicate the IMSBS consistently produces the highest-quality solutions over the baseline beam search. In cases where shorter expected solutions are anticipated, frequently transitioning between different feasible subspaces is more effective. This is in contrast to situations with a small sequence number, where a larger beam width and a greater number of IMSBS iterations are necessary to obtain high-quality solutions.  %Concerning exact solving, we  compared three dynamic programming approaches from the literature with a newly designed ILP model. The results confirmed the DP variants solved all instances in the shortest amount of time, while the ILP approach was limited to smaller instances.  Additionally, IMSBS approach has frequently achieved close-to-optimum results for the two-input-sequence problem case, while requiring comparable or less computation time than the baseline beam search.
 Future research includes designing more sophisticated heuristics which incorporate gap constraints into scoring, optimizing IMSBS to efficiently reuse previously explored regions, % of the state space, 
 and testing the approaches on real-world biological instances. % and scaling experiments up to longer sequences. 
 \\
 \textit{\textbf{Acknowledgments.}} This publication is co-funded by the European Union’s Horizon Europe research and innovation program under the Marie Sklodowska-Curie COFUND Postdoctoral Programme (grant agreement No.~101081355 -- SMASH), and by the Republic of Slovenia and the European Union through the European Regional Development Fund. Views and opinions expressed are those of the authors only and do not necessarily reflect those of the European Union or the European Research Executive Agency (REA). Neither the European Union nor the REA can be held responsible for them.

	\bibliographystyle{splncs04}
	\bibliography{bib}

\begin{thebibliography}{10}
\providecommand{\url}[1]{\texttt{#1}}
\providecommand{\urlprefix}{URL }
\providecommand{\doi}[1]{https://doi.org/#1}

\bibitem{adamson2023longest}
Adamson, D., Kosche, M., Ko{\ss}, T., Manea, F., Siemer, S.: Longest common
  subsequence with gap constraints. In: International Conference on
  Combinatorics on Words. pp. 60--76. Springer (2023)

\bibitem{adi2010repetition}
Adi, S.S., Braga, M.D., Fernandes, C.G., Ferreira, C.E., Martinez, F.V., Sagot,
  M.F., Stefanes, M.A., Tjandraatmadja, C., Wakabayashi, Y.: Repetition-free
  longest common subsequence. Discrete Applied Mathematics  \textbf{158}(12),
  1315--1324 (2010)

\bibitem{bergroth2000survey}
Bergroth, L., Hakonen, H., Raita, T.: A survey of longest common subsequence
  algorithms. In: Proceedings Seventh International Symposium on String
  Processing and Information Retrieval. SPIRE 2000. pp. 39--48. IEEE (2000)

\bibitem{blum2016metaheuristics}
Blum, C., Festa, P.: Metaheuristics for String Problems in Bio-informatics,
  vol.~6. John Wiley \& Sons (2016)

\bibitem{DJUKANOVIC2020106499}
Djukanovic, M., Raidl, G., Blum, C.: Finding longest common subsequences: New
  anytime {A*} search results. Applied Soft Computing  \textbf{95},  106499
  (2020)

\bibitem{farhana2015constrained}
Farhana, E., Rahman, M.S.: Constrained sequence analysis algorithms in
  computational biology. Information Sciences  \textbf{295},  247--257 (2015)

\bibitem{jiang2004longest}
Jiang, T., Lin, G., Ma, B., Zhang, K.: The longest common subsequence problem
  for arc-annotated sequences. Journal of Discrete Algorithms  \textbf{2}(2),
  257--270 (2004)

\bibitem{lainscsek2015delay}
Lainscsek, C., Sejnowski, T.J.: Delay differential analysis of time series.
  Neural computation  \textbf{27}(3),  594--614 (2015)

\bibitem{mousavi2012improved}
Mousavi, S.R., Tabataba, F.: An improved algorithm for the longest common
  subsequence problem. Computers \& Operations Research  \textbf{39}(3),
  512--520 (2012)

\bibitem{penga2011longest}
Penga, Y.H., Yangb, C.B.: The longest common subsequence problem with variable
  gapped constraints. In: Proceedings of the 28th Workshop on Combinatorial
  Mathematics and Computation Theory, Penghu, Taiwan. pp. 17--23 (2011)

\end{thebibliography}


\begin{thebibliography}{99}
 	
 	\bibitem{blum2016metaheuristics}
 	C.~Blum and P.~Festa,
 	\newblock {\em Metaheuristics for String Problems in Bio-informatics},
 	\newblock Vol.~6, John Wiley \& Sons, 2016.
 	
 	\bibitem{penga2011longest}
 	Y.-H.~Penga and C.-B.~Yangb,
 	\newblock The Longest Common Subsequence Problem with Variable Gapped Constraints,
 	\newblock In {\em Proceedings of the 28th Workshop on Combinatorial Mathematics and Computation Theory, Penghu, Taiwan}, pp.~17--23, 2011.
 	
 \end{thebibliography}
 \newpage
 \appendix
 \section{ILP model for the fixed $m=2$ VGLCS Problem}\label{app:ilp}
 
 For the case of the VGLCS problem with $m=2$, in this section, we design an
 \emph{integer programming model}, representing a methodological approach that differs from
 most existing approaches in the literature, which are predominantly based on dynamic
 programming (DP). This is a proof-of-concept showing the structural difficulty of this approach and a baseline modeling contribution when solving the fixed version of the tackled problem. Shortly, the purpose of the ILP model is not computational competitiveness but structural modeling and benchmarking. \\
 
 To this end, let:
 \begin{itemize}
 	\item $M = \{ (i,j) \mid s_1[i] = s_2[j] \}$ denotes the set of all positions where the characters match.
 \end{itemize}
 
 For each $(i,j) \in M$, we define a binary variable:
 \[
 x_{i,j} =
 \begin{cases}
 	1, & \textrm{if the pair } (i,j) \textrm{ is selected as part of the solution}, \\
 	0, & \textrm{otherwise}.
 \end{cases}
 \]
 
 We introduce additional binary variables $s_{i,j}$ indicating whether the pair $(i,j)$ represents
 the \emph{starting position of the subsequence}. \\
 
 \textit{Objective function}. We maximize the number of selected admissible matches:
 \[
 \max \sum_{(i,j) \in M} x_{i,j}
 \]
 
 \textit{Constraints}. The following constraints are imposed:
 
 \begin{enumerate}
 	\item \textrm{Predecessors (variable gap).}  
 	For each $(i,j) \in M$, we define the set of valid predecessors:
 	\[
 	\texttt{Pred}(i,j) =
 	\{ (i',j') \in M \mid i' < i,\ j' < j,\ i - i' \leq G_{s_1}[i] + 1,\ j - j' \leq G_{s_2}[j] + 1 \}
 	\]
 	The activation constraint is given by:
 	\[
 	x_{i,j} \leq \sum_{(i',j') \in \textrm{Pred}(i,j)} x_{i',j'} + s_{i,j}
 	\]
 	
 	\item \textit{Starting position.}  
 	At most one starting pair is allowed:
 	\[
 	\sum_{(i,j) \in M} s_{i,j} \leq 1
 	\]
 	
 	\item \textit{Conflict constraints (character ordering).}  
 	Two distinct pairs $(i,j)$ and $(i',j')$ are in conflict if they violate the character order:
 	\[
 	\textrm{If } (i \leq i' \textrm{ and } j \geq j') \textrm{ or }
 	(i \geq i' \textrm{ and } j \leq j'), \textrm{ then:}
 	\quad x_{i,j} + x_{i',j'} \leq 1
 	\]
 	\item \textit{Auxiliary constraints}: $s_{i, j}=1 \Rightarrow x_{i',j'}=0, \forall (i' \leq i-1, j' \leq j-1)$. The constraints are expressed in the following form: 
 	$$ \sum_{(i', j') \mid i'\leq i \wedge j'\leq j} x_{i' j'} + s_{i, j} \leq 1, \forall (i, j) \in M$$
 \end{enumerate}

 \textit{Domains of the variables are given by}
 \[
 x_{i,j},\ s_{i,j} \in \{0,1\}, \quad \forall (i,j) \in M.
 \]
 
 The model is formulated to maximize the number of valid matches forming a common subsequence,
 while respecting the ordering of elements and the maximum allowed gaps between consecutive
 elements in both sequences.
 
 \section{Pseudocodes of the IMSBS approach}
 
 \begin{algorithm}[H]
 	\caption{Rooted Beam Search}
 	\label{alg:bs}
 	\begin{algorithmic}[1]
 		
 		\Function{RootedBeamSearch}{$R, S, \{G_i\}_{i=1}^m, \beta, h$}
 		\Comment{$r\in R$: root nodes, $S=\{s_1,\dots,s_m\}$, gap constraints $G_i$, beam width $\beta$, heuristic $h$}
 		
 		\State $s_{\text{best}} \gets \varepsilon$
 		\State $B \gets \emptyset$ \Comment{Starting beam}
 		%the first extension: TODO
 		\ForAll{$v \in R$} \Comment{starting match extensions}
 		\State $B \gets B \cup \{ (p^{L,v}+\textbf{1}, 1) \}$
 		\EndFor
 		\State $V_{\text{complete}} \gets \emptyset$
 		
 		\While{$B \neq \emptyset$}
 		\State $Children \gets \emptyset$
 		
 		\ForAll{$v \in B$}
 		\If{$v$ is complete}
 		\State $V_{\text{complete}} \gets V_{\text{complete}} \cup \{v\}$
 		
 		\If{$v.l^v > |s_{\text{best}}|$}
 		\State $s^v \gets$ Extract partial solution associated with $v$
 		\State $s_{\text{best}} \gets s^v$
 		\EndIf
 		
 		\State \textbf{continue}
 		\EndIf
 		
 		\State $C_v \gets$ \Call{Expand}{$v, S, \{G_i\}_{i=1}^m$}
 		\State $Children \gets Children \cup C_v$
 		\EndFor
 		
 		\State \Call{Sort}{$Children, h$} \Comment{decreasing order}
 		\State $B \gets Children[ \colon \beta]$
 		\EndWhile
 		
 		\State \Return $s_{\text{best}}, V_{\text{complete}}$
 		\EndFunction
 	\end{algorithmic}
 \end{algorithm}

 %pseudocode:
 \begin{algorithm}[H]
 	\caption{Iterative Multi-source Beam Search (IMSBS) Framework }
 	\label{alg:imbs}
 	\begin{algorithmic}[1]
 		\Require Sequences $S=\{s_1, \dots, s_m\}$; Gap functions $G_{s_i}(\cdot),i=1,\ldots,m$; beam width $\beta>0$; ${sources\_from\_R}>0$; $beam\_iters>0$, heuristics $h, h'$, time limit $t_{{lim}}$
 		\Ensure Approximate common subsequence $s_{\textrm{best}}$ which fulfills all gap constraints
 		
 		\State Initialize $\mathcal{R} \gets \{ (\mathbf{1},0) \}$ \Comment{A set of root nodes}

 		\ForAll{ $a \in \Sigma$} \Comment{Find closest matching positions for each letter}
 		\State $\textbf{p}^w_{a} = ({p}^w_{i, a})_{i=1}^m \gets$ minimum position  $p^w_{i, a} \geq 1$ and $s_i[p^w_{i, a}] = a$ (ignore all gap constraints, remove dominated root positions)
 		\State $R \gets R \cup \{ (p^w_a, 0)\}$
 		\EndFor
 		\State $iter \gets 0$
 		\State $s_{\textrm{best}} \gets  \varepsilon$
 		
 		\While{$\mathcal{R} \neq \emptyset \wedge iter < beam\_iters  \wedge t_{lim}$ not exceeded }
 		\State Select $\mathcal{L} \subseteq \mathcal{R}$  ${sources\_from\_R}$  and remove them from $\mathcal{R}$ \Comment{ $\mathcal{R}$ prioritized acc. to $h'$}
 		%\Comment{$\mathcal{R}$ can be a pr. queue ordered by heuristic $h'$}
 		\ForAll{$w \in \mathcal{L}$}
 		\State $s_{bwrd}^w, \_ \gets$ 
 		\textsc{RootedBeamSearch}(  
 		$(\{ (|s_i| - p^w_i)_{i=1}^m, 0\}, S^{rev}, G^{rev}, \beta', h')$
 		\Comment{Refine each $w \in \mathcal{L}$}
 		\State  $l^w \gets |(s_{bwrd}^w)^{rev}|-1$  \Comment{Update $l^w$, assign the reverse of $s_{bwrd}^w$ without the last letter  to $w$}
 		\EndFor
 		\State $s_{b}, V_{complete}
 		\gets$ 
 		\textsc{RootedBeamSearch}($\mathcal{L}, S, G, \beta, h)$ 
 		\If{$|s_b| > s_{best}$}
 		\State $s_{best} \gets s_b$
 		\EndIf
 		
 		\ForAll{$v \in V_{complete}$}
 		
 		\ForAll{ $a \in \Sigma$} \Comment{Find closest matching positions for each letter}
 		\State $\textbf{p}^w_{a}= ({p}^w_{i, a})_{i=1}^m \gets$ minimum position  $p^w_{i, a} \geq {p}_i^v$ and $s_i[p^w_{i, a}] = a$ (ignore all gap constraints)
 		%\If{$|s_{\textrm{best}}| < l^w + \min_{i=1,\dots,m} \{ |s_i| - p^w_i + 1 \}$}    \Comment{cut-off}
 		
 		\State $\mathcal{R} \gets$ $\mathcal{R} \cup \{r^w = (\textbf{p}^w_a, 0)\}$   (if has not previously been  added) \Comment{Update $\mathcal{R}$}
 		% \EndIf
 		\EndFor
 		\EndFor
 		
 		\State $iter \gets iter + 1$
 		\EndWhile
 		
 		\State \Return $s_{\textrm{best}}$
 	\end{algorithmic}
 \end{algorithm}

 \begin{figure}[!ht]
 	\centering
 	\scalebox{0.70}{
 		\begin{tikzpicture}[
 			node distance=0.7cm and 1.2cm,
 			every node/.style={font=\small},
 			block/.style={
 				rectangle,
 				draw,
 				rounded corners,
 				align=center,
 				minimum width=2.0cm,
 				minimum height=0.6cm,
 				fill=green!10
 			},
 			decision/.style={
 				diamond,
 				draw,
 				align=center,
 				aspect=2,
 				fill=green!5
 			},
 			arrow/.style={
 				->,
 				thick,
 				>=Latex,
 				shorten >=2pt,
 				shorten <=2pt
 			}
 			]
 			
 			% Nodes
 			\node[block] (start) {Start};
 			
 			\node[block, below=of start] (init) {Initialize \textbf{global queue of roots} $\mathcal{R}=\{\}$\\
 				$s_{\text{best}} \leftarrow \varepsilon$};
 			
 			\node[block, below=of init] (expandR) {Initial generation of sources:\\ Determine the \textbf{closest matches} for each $a \in \Sigma$ from the top positions};
 			
 			\node[decision, below=of expandR] (while) {A termination criterion not met};
 			
 			\node[block, below=of while] (selectL) {Select subset $\mathcal{L} \subseteq \mathcal{R}$\\
 				(priority by $h'$)};
 			
 			\node[block, below=of selectL] (backward) {\textbf{Refinement} of $v \in \mathcal{L}$\\
 				(Rooted BS on reversed prefixes of $S$ w.r.t.\ $p^{L,v}$)};
 			
 			\node[block, below=of backward] (forward) {Rooted (\textbf{forward}) BS using $\mathcal{L}$ for its initial beam};
 			
 			\node[block, below=of forward] (updateBest) {Update best solution $s_{\text{best}}?$};
 			
 			\node[block, below=of updateBest] (expandNew) {\textbf{Generate new candidate roots} from comple states\\
 				(ignoring gaps, closest match detection), \\ update $\mathcal{R}$};
 			
 			\node[block, below=of expandNew] (end) {Return $s_{\text{best}}$};
 			
 			% Arrows (main flow)
 			\draw[arrow] (start) -- (init);
 			\draw[arrow] (init) -- (expandR);
 			\draw[arrow] (expandR) -- (while);
 			
 			\draw[arrow] (while) -- node[right] {yes} (selectL);
 			\draw[arrow] (selectL) -- (backward);
 			\draw[arrow] (backward) -- (forward);
 			\draw[arrow] (forward) -- (updateBest);
 			\draw[arrow] (updateBest) -- (expandNew);
 			
 			% Explicit loop-back arrow (FIXED)
 			\draw[arrow]
 			(expandNew.east) -- ++(2.7,0)
 			|- (while.east);
 			
 			% Exit arrow
 			\draw[arrow]
 			(while.west) -- ++(-1.8,0)
 			node[left] {no}
 			|- (end.west);		
 	\end{tikzpicture}}
 	
 	\caption{Simplified workflow of the Iterative Multi-source Beam Search (IMSBS).}
 	\label{fig:imsbs_workflow}
 \end{figure}
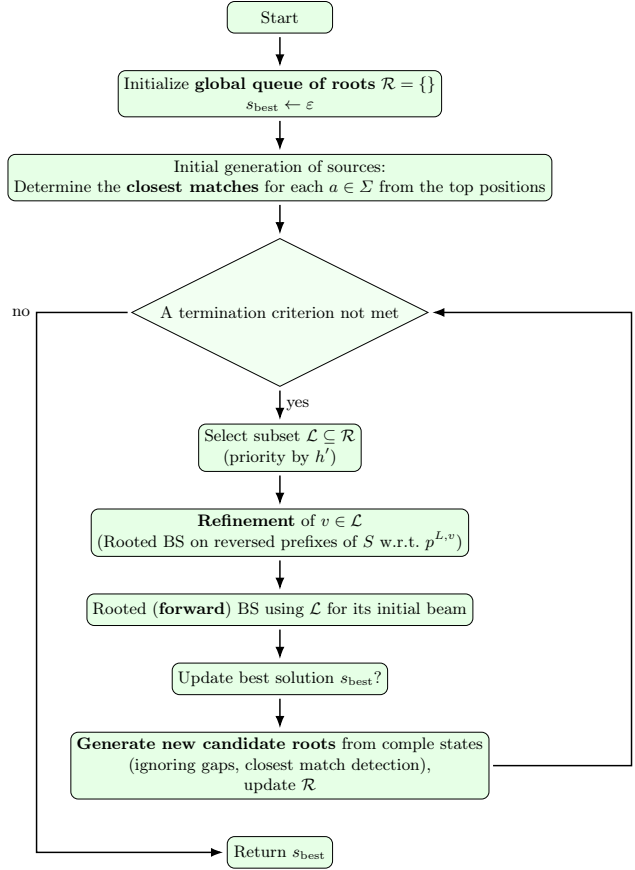

 \section{Experimental Evaluation: Tuning \& Numerical Results}
 
 \begin{figure}[H]
 	\centering
 	\begin{subfigure}[t]{0.48\textwidth}
 		\centering
 		\includegraphics[width=\linewidth, height=120pt]{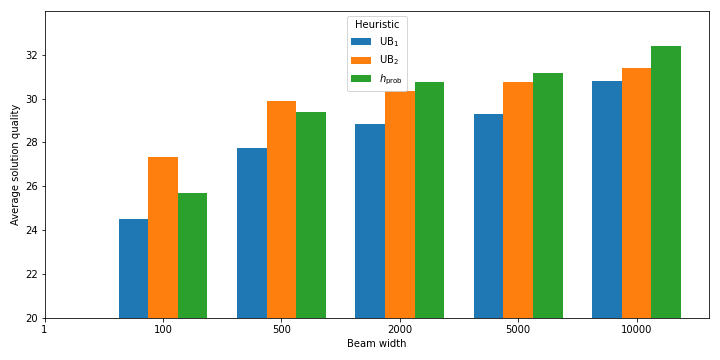}
 		\caption{Avg. quality over all instances of the \textsc{Random} benchmark suite.}
 		\label{fig:tune_bs}
 	\end{subfigure}
 	~
 	\begin{subfigure}[t]{0.48\textwidth}
 		\centering
 		\includegraphics[width=\linewidth, height=120pt]{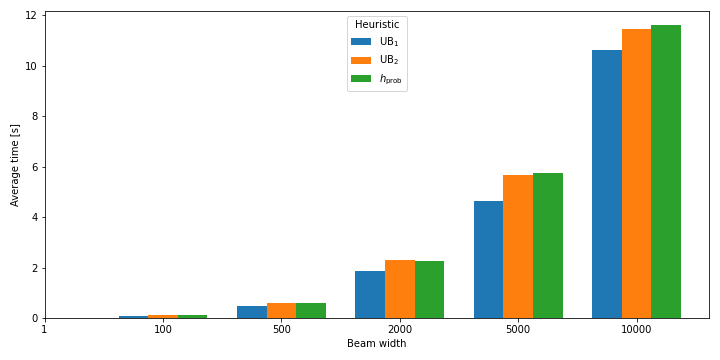}
 		\caption{Avg. runtime over all instances of the \textsc{Random} benchmark suite.}
 		\label{fig:tune_bs_time}
 	\end{subfigure}
 	\caption{Beam search tuning results on the \textsc{Random} benchmark suite.}
 	\label{fig:bs_tuning}
 \end{figure}
 
 \begin{figure}[!h]
 	\centering
 	\begin{minipage}{0.48\textwidth}
 		\centering
 		\includegraphics[width=\linewidth,height=120pt]{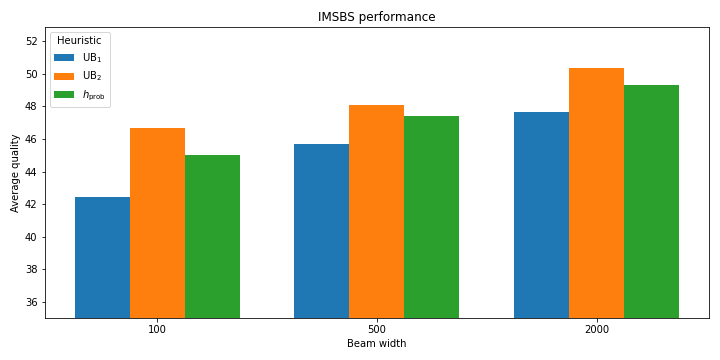}
 		\caption*{(a) Avg. solution quality of different IMSBS settings over all instances of the \textsc{Random} benchmark suite.}
 		\label{fig:tune_imsbs}
 	\end{minipage}
 	~
 	\begin{minipage}{0.48\textwidth}
 		\centering
 		\includegraphics[width=\linewidth,height=120pt]{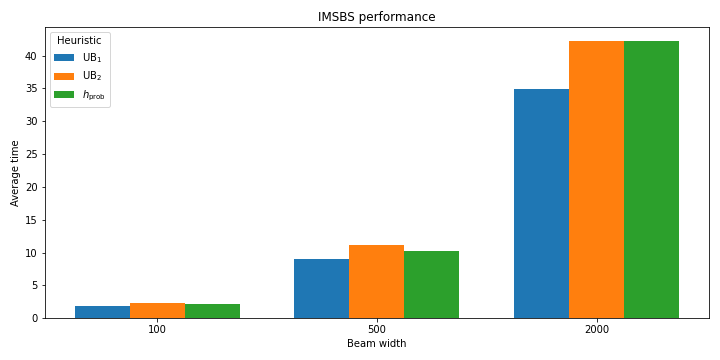}
 		\caption*{(b) Avg. runtime of different IMSBS settings over all instances of the \textsc{Random} benchmark suite.}
 		\label{fig:tune_imsbs_time}
 	\end{minipage}
 	\caption{IMSBS parameter tuning results on the \textsc{Random} benchmark suite.}
 	\label{fig:imsbs_tuning}
 \end{figure}

 Numerical results for all three heuristic derivatives are shown in Table~\ref{tab:numerical_results_general_m}.   organized as follows. 
 First three columns describe the instance characteristics: the number of sequences ($m$), the sequence length ($n$), and the alphabet size ($|\Sigma|$).  The remaining columns report the performance metrics of three heuristic approaches: \textsc{Bs}, \textsc{Imsbs-greedy}, and \textsc{Imsbs} respectively, each by its own block where two performance indicators are reported: the average objective value ($\overline{obj}$) and the average running time in seconds ($\overline{t}$), computed over 10 instances per group. 
 
 \begin{table}[!ht]
 	\caption{Numerical results   on the \textsc{Random} benchmark suite. }\label{tab:numerical_results_general_m}
 	\centering
 	\scalebox{0.80}{
 		\begin{tabular}{lll|rr|rr|rr}
 			\hline
 			\multicolumn{3}{c}{Inst.} & \multicolumn{2}{c}{\textsc{Bs}} & \multicolumn{2}{c}{\textsc{Imsbs-Greedy}} & \multicolumn{2}{c}{\textsc{Imsbs}} \\
 			\cmidrule(lrr){1-3} \cmidrule(lr){4-5} \cmidrule(lr){6-7} \cmidrule(lr){8-9} \\ 
 			$m$ & $n$ & $|\Sigma|$ & $\overline{obj}$ & $\overline{t}[s]$ & $\overline{obj}$ & $\overline{t}[s]$ & $\overline{obj}$ & $\overline{t}[s]$ \\
 			\hline
 			\midrule
 			2 &       50 &         2 &            33.6 &         0.02 &                     33.1 &                  0.00 &         \textbf{37.7} &      0.06 \\
 			2 &       50 &         4 &            30.1 &         0.98 &                     27.7 &                  0.00 &         \textbf{30.1} &      0.16 \\
 			2 &      100 &         2 &            48.9 &         2.07 &                     64.5 &                  0.01 &         \textbf{72.8} &      0.94 \\
 			2 &      100 &         4 &            \textbf{62.1} &        11.19 &                     56.9 &                  0.01 &         61.6 &      0.91 \\
 			2 &      200 &         2 &            99.1 &        18.56 &                     95.5 &                  0.02 &        \textbf{136.4} &      6.21 \\
 			2 &      200 &         4 &           120.5 &        38.15 &                    116.1 &                  0.05 &        \textbf{124.9} &      6.58 \\
 			2 &      500 &         2 &            65.3 &        23.27 &                    153.6 &                  0.07 &        \textbf{265.7} &    119.75 \\
 			2 &      500 &         4 &           214.6 &       163.69 &                    227.7 &                  0.12 &        \textbf{294.4} &     60.49 \\ \hline
 			3 &       50 &         2 &            17.5 &         0.03 &                     27.2 &                  0.00 &         \textbf{31.2} &      0.14 \\
 			3 &       50 &         4 &            21.7 &         0.18 &                     21.5 &                  0.00 &         \textbf{22.9} &      0.19 \\
 			3 &      100 &         2 &            19.7 &         0.06 &                     41.8 &                  0.01 &         \textbf{58.5} &      3.15 \\
 			3 &      100 &         4 &            34.1 &         2.35 &                     43.4 &                  0.03 &         \textbf{48.4} &      5.58 \\
 			3 &      200 &         2 &            15.2 &         0.16 &                     63.6 &                  0.02 &         \textbf{90.0} &     22.48 \\
 			3 &      200 &         4 &            85.3 &        23.45 &                     77.2 &                  0.08 &         \textbf{97.1} &     72.25 \\
 			3 &      500 &         2 &            12.6 &         0.10 &                     69.9 &                  0.07 &        \textbf{102.9} &     53.56 \\
 			3 &      500 &         4 &            90.7 &        86.35 &                    104.2 &                  0.29 &        \textbf{187.7} &    412.27 \\ \hline
 			5 &       50 &         2 &             4.8 &         0.00 &                     14.9 &                  0.00 &         \textbf{20.0} &      0.36 \\
 			5 &       50 &         4 &             8.9 &         0.01 &                     13.6 &                  0.06 &         \textbf{15.3} &      0.79 \\
 			5 &      100 &         2 &             6.3 &         0.01 &                     17.7 &                  0.01 &         \textbf{22.4} &      0.57 \\
 			5 &      100 &         4 &             5.3 &         0.01 &                     \textbf{23.2} &                 10.85 &         22.1 &      1.44 \\
 			5 &      200 &         2 &             5.3 &         0.01 &                     21.6 &                  0.03 &         \textbf{26.6} &      1.07 \\
 			5 &      200 &         4 &             6.4 &         0.02 &                     \textbf{32.5} &                604.11 &         {25.7} &      2.10 \\
 			5 &      500 &         2 &             5.9 &         0.10 &                     25.5 &                  0.14 &         \textbf{27.9} &      3.22 \\
 			5 &      500 &         4 &             6.8 &         0.10 &                     \textbf{43.6} &               1341.25 &         26.9 &      3.52 \\ \hline
 			10 &       50 &         2 &             1.7 &         0.00 &                      8.8 &                  2.28 &          \textbf{9.1} &      0.47 \\ 
 			10 &       50 &         4 &             1.9 &         0.00 &                      \textbf{7.0} &                508.36 &          6.1 &      1.46 \\
 			10 &      100 &         2 &             1.1 &         0.00 &                     \textbf{14.0} &               1421.10 &          8.6 &      0.54 \\
 			10 &      100 &         4 &             2.2 &         0.01 &                      \textbf{8.9} &               1800.45 &          {6.3} &      1.51 \\
 			10 &      200 &         2 &             2.5 &         0.01 &                     \textbf{13.2} &               1710.49 &         10.3 &      0.77 \\
 			10 &      200 &         4 &             2.2 &         0.02 &                      \textbf{7.9} &               1800.54 &          6.1 &      1.74 \\
 			10 &      500 &         2 &             1.8 &         0.08 &                     \textbf{13.8} &               1611.70 &          {9.5} &      1.53 \\
 			10 &      500 &         4 &             1.9 &         0.09 &                      \textbf{8.2} &               1800.46 &          6.1 &      2.37 \\
 			\hline \hline
 			\textbf{Avg.} &  &  &  32.38 & 10.97 & 46.82 & 394.14 &  \textbf{59.73} & 24.63  \\ \hline \hline
 	\end{tabular}}
 \end{table}
 
 \section{Numerical results for the $m=2$ case}
 
 In this section, we compare the results of approaches from the literature that are specialized for the case $m=2$. The competitors are listed below and are described in detail in~\cite{penga2011longest}.

\begin{itemize}
	\item \textsc{Dp}: the basic dynamic programming algorithm;
	\item \textsc{Dp}-1: an advanced dynamic programming approach that uses Incremental Suffix Maximum Queries (ISMQ) with \texttt{Col} and \texttt{All} matrices to accelerate the basic DP algorithm;
	\item \textsc{Dp}-2: an enhanced dynamic programming algorithm that handles ISMQ using a \textit{dequeue} data structure for further speedup. All three DP approaches are from \cite{penga2011longest}. Note that the original paper proposes answering ISMQ using Union-Find operations; however, due to a lack of implementation details, we decided to employ a dequeue-based approach in our implementation.
	\item \textsc{Ilp}: an integer linear programming method proposed in this work (see Appendix~\ref{app:ilp}), motivated by the ILP model for the LCS problem~\cite{blum2016metaheuristics}.
\end{itemize}

Each algorithm was allocated 30 minutes of execution time. The \textsc{Ilp} model was solved using the general-purpose solver \textsc{Cplex} version 22.1.

The results are reported in Table~\ref{tab:results-2d-literature}, which is organized as follows. The first three columns describe the characteristics of the instance groups, over which the results are averaged (10 instances per group). The remaining columns report the performance of the four algorithms: \textsc{Dp}, \textsc{Dp}-1, \textsc{Dp}-2, and \textsc{Ilp}. For each approach, both the average solution quality and the average execution time are provided across 10 instances.

\begin{table}[H]
	\caption{Results on the \textsc{Random} benchmark set for $m=2$: the exact approaches from the literature.}\label{tab:results-2d-literature}
	\centering
	\scalebox{0.9}{
		\begin{tabular}{lll|lr|rr|rr|rr}
			\hline
			\multicolumn{3}{c}{Inst.} & \multicolumn{2}{c}{\textsc{Dp}}  &
			\multicolumn{2}{c}{\textsc{Dp-1}} & \multicolumn{2}{c}{\textsc{Dp-2}} & 
			\multicolumn{2}{c}{\textsc{Ilp}}  \\
			\cmidrule(lrr){1-3} \cmidrule(lr){4-5}
			\cmidrule(lr){6-7} \cmidrule(lr){8-9}
			\cmidrule(lr){10-11} 	\\ 
			$m$ & $n$ & $|\Sigma|$ & $\overline{obj}$ & $\overline{t}[s]$ & $\overline{obj}$  & $\overline{t}[s]$ &$\overline{obj}$  & $\overline{t}[s]$ & $\overline{obj}$  & $\overline{t}[s]$ \\
			\hline
			2 &   50 &         2 &              38.1 &          0.01 & 38.1 &          \textbf{0.01} &              38.1 &          0.01  & 38.1 & 168.3 \\
			2 &   50 &         4 &              30.3 &          \textbf{0.01} &              30.3 &          0.02 &              30.3 &          0.02 & 30.3  & 28.0 \\ \hline
			2 &  100 &         2 &              77.4 &          0.1 &              77.4 &          \textbf{0.03} &              77.4 &          0.05 & -- & -- \\
			2 &  100 &         4 &              62.3 &          0.07 &              62.3 &         \textbf{0.06} &              62.3 &          0.09 & 0.00 & 1800.0 \\ \hline
			
			2 &  200 &         2 &             156.4 &          0.75 &             156.4 &          \textbf{0.13} &             156.4 &          0.16  & -- & -- \\
			2 &  200 &         4 &             127.2 &          0.59 &             127.2 &          \textbf{0.25} &             127.2 &          0.32  & -- & -- \\ \hline
			
			2 &  500 &         2 &             395.9 &         13.57 &             395.9 &          \textbf{0.84}  &             395.9 &          1.05 & -- & -- \\
			2 &  500 &         4 &             317.2 &         10.18 &             317.2 &          \textbf{1.70} & 317.2 &  2.1 & -- &-- \\    
			\hline \hline
	\end{tabular}}
	
\end{table}

Based on the numerical results reported in Table~\ref{tab:results-2d-literature}, the following conclusions can be drawn:

\begin{itemize}
	\item All three dynamic programming approaches prove to be highly efficient, successfully solving all 80 problem instances to optimality.
	\item The runtimes required by the DP methods to obtain optimal solutions are relatively short, in most cases remaining below 10 seconds.
	\item The \textsc{ILP} approach is capable of solving only the smallest instances with $n=50$. Its execution times are approximately two orders of magnitude higher than those of the DP approaches. For larger instances, particularly those with $n=200$, the ILP model fails to solve even the initial (root) relaxation.
\end{itemize}

\begin{figure}[htbp]
	\centering
	\begin{minipage}{0.90\textwidth}
		\includegraphics[width=\linewidth,height=140pt]{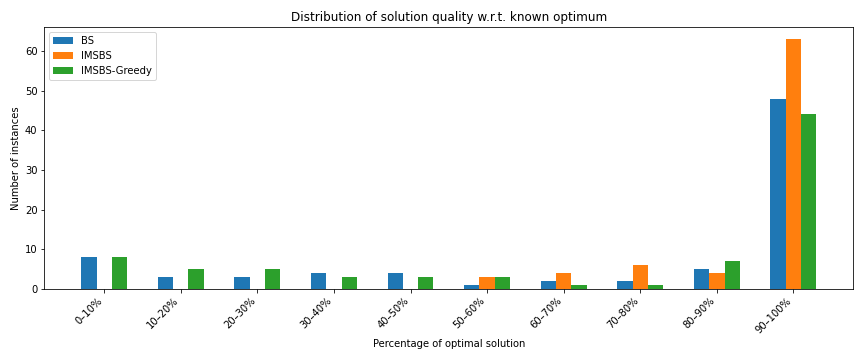} 
		\caption*{(a) Percentage distribution of relative solution quality for each heuristic approach with respect to the optimal solutions.} 
	\end{minipage}
	\hfill
	\begin{minipage}{0.90\textwidth}
		\centering
		\includegraphics[width=\linewidth,height=140pt]{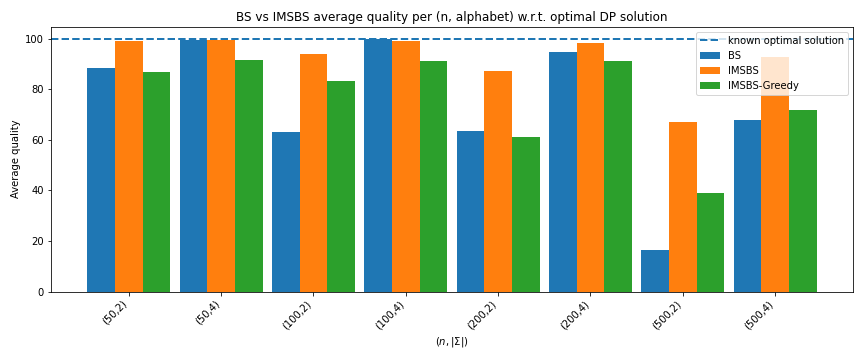}
		\caption*{(b) Relative solution quality achieved by the heuristic approaches compared to the optimal solutions.}
	\end{minipage}
	\caption{Number of instances for which \textsc{Bs} and \textsc{Imsbs} achieve a specific relative ratio of the obtained solution quality with respect to the known optimal solutions.}
	\label{fig:optimal-vs-heuristic}
\end{figure}

Concerning the efficiency of \textsc{Bs}, \textsc{Imsbs-Greedy}, and \textsc{Imsbs} on the instances solved optimally by dynamic programming ($m=2$), Figure~\ref{fig:optimal-vs-heuristic} presents the distribution of instances (y-axis) attaining specific ratios of solution quality relative to the known optimal solutions (x-axis).

In particular, \textsc{Imsbs} achieves near-optimal solutions in 63 out of 80 problem instances, whereas the corresponding numbers are significantly lower for the other two heuristic approaches. For instance, with $|\Sigma|=4$ (and $m=2$), the difference in solution quality is slightly in favor of \textsc{Imsbs}. However, for instances with a small alphabet ($|\Sigma|=2$), \textsc{Imsbs} clearly outperforms the other two approaches.

Nevertheless, for the group of instances with $n=500$ and $|\Sigma|=2$, there remains room for further improvement. This behavior can be explained by the known weakness of LCS-based heuristics on instances characterized by small alphabet sizes and long sequences; therefore, alternative strategies should be explored, as discussed in the future work section.

\end{document}